%% file: paper.tex
\documentclass[10pt, final, conference]{IEEEtran}

\usepackage[english]{babel}
\usepackage{csquotes}

\usepackage[style=ieee, sorting=none, backend=biber]{biblatex}
\usepackage{amsmath}
\usepackage{amsfonts}
\usepackage{amssymb}

\usepackage{etoolbox}
\AtBeginEnvironment{equation*}{\small}

\usepackage{balance}

\usepackage{graphicx}
\usepackage{float}

\usepackage{pgfplots}
\pgfplotsset{compat=newest}

\usepackage{pgf-pie}

\newcommand*{\TitleFont}{%
      \usefont{\encodingdefault}{\rmdefault}{b}{n}%
      \fontsize{20}{20}%
      \selectfont}

\usepackage{tabularx}
\usepackage{booktabs}

\usepackage{hyperref}
\hypersetup{
	hidelinks,
	pdftitle={A Framework for Genetic Algorithms Based on Hadoop},
	pdfauthor={Filomena Ferrucci, Tahar M. Kechadi, Pasquale Salza, Federica Sarro},
	pdfkeywords={Parallel Genetic Algorithms, Hadoop, MapReduce, Framework}
}

\usepackage{newclude}

\begin{document}

	\include*{title}
	\include*{abstract}
	\include*{keywords}

	\include*{introduction}
	\include*{background_and_related_work}
	\include*{design_and_development}
	\include*{preliminary_analysis}
	\include*{conclusions_and_future_work}

	\include*{bibliography}

\end{document}

%% file: title.tex
\title{\TitleFont A Framework for Genetic Algorithms Based on Hadoop}

\author{
	\IEEEauthorblockN{
		Filomena Ferrucci\IEEEauthorrefmark{1},
		M-Tahar Kechadi\IEEEauthorrefmark{2},
		Pasquale Salza\IEEEauthorrefmark{1},
		Federica Sarro\IEEEauthorrefmark{3}
	} \\
	\IEEEauthorblockA{
		\IEEEauthorrefmark{1}
		Department of Management and Information Technology, DISTRA (MIT) \\
		Universit\`{a} degli Studi di Salerno, Italy \\
		e-mail: \href{mailto:fferrucci@unisa.it}{\nolinkurl{fferrucci@unisa.it}}
	} \\
	\IEEEauthorblockA{
		\IEEEauthorrefmark{2}
		School of Computer Science and Informatics \\
		University College Dublin, Ireland \\
		e-mail: \href{mailto:tahar.kechadi@ucd.ie}{\nolinkurl{tahar.kechadi@ucd.ie}}
	} \\
	\IEEEauthorblockA{
		\IEEEauthorrefmark{3}
		Department of Computer Science, CREST \\
		University College London, United Kingdom \\
		e-mail: \href{mailto:f.sarro@cs.ucl.ac.uk}{\nolinkurl{f.sarro@cs.ucl.ac.uk}}
	}
}

\maketitle

\thispagestyle{plain}
\pagestyle{plain}

%% file: abstract.tex
\begin{abstract}

	\textit{Genetic Algorithms} (\textit{GAs}) are powerful metaheuristic techniques mostly used in many real-world applications. The sequential execution of \textit{GAs} requires considerable computational power both in time and resources. Nevertheless, \textit{GAs} are naturally parallel and accessing a parallel platform such as \textit{Cloud} is easy and cheap. \textit{Apache Hadoop} is one of the common services that can be used for parallel applications. However, using \textit{Hadoop} to develop a parallel version of \textit{GAs} is not simple without facing its inner workings. Even though some sequential frameworks for \textit{GAs} already exist, there is no framework supporting the development of \textit{GA} applications that can be executed in parallel. In this paper is described a framework for parallel \textit{GAs} on the \textit{Hadoop} platform, following the paradigm of \textit{MapReduce}. The main purpose of this framework is to allow the user to focus on the aspects of \textit{GA} that are specific to the problem to be addressed, being sure that this task is going to be correctly executed on the \textit{Cloud} with a good performance. The framework has been also exploited to develop an application for \textit{Feature Subset Selection} problem. A preliminary analysis of the performance of the developed \textit{GA} application has been performed using three datasets and shown very promising performance.
	
\end{abstract}

%% file: keywords.tex

%% file: introduction.tex
\section{Introduction}

	For  problems  with non  polynomial  complexity  the search  for  an  optimum solution  is considered to be a mission impossible, as it involves huge resources and execution time. Metaheuristic techniques, such as \textit{Genetic Algorithms} (\textit{GAs}), constitute the best alternative to find  near-optimal solutions for such problems within a reasonable execution time and limited resources.

	\textit{GAs} approach mimics the biological process of reproduction. It starts with an initial population of individuals, each of which is represented by a \textit{chromosome}. The \textit{GA} iterates by executing some common operations on the selected individuals, which are \textit{crossover} and \textit{mutation}. The population tends to improve	its individuals by keeping the strongest individuals and rejecting the weakest ones. However, one of the main drawbacks of the technique is how to model the real-world problem into a genetic one. The  model, usually  called coding, is far from being straightforward. In order to model an original optimisation problem into a genetic one, some main functions of \textit{GAs} need to be defined properly. The \textit{fitness function} should correspond to objective function of the original problem. This function will be used to evaluate the individuals. Usually the individuals with good fitness will be selected for next generation.
	
	Often, \textit{GAs} are executed on single machines as sequential programs. However, the main principle behind these algorithms is not really sequential, as it is possible to select more than two individuals for reproduction, use more than one population, and execute the operators in parallel. Parallel systems are becoming commonplace mainly with the increasing popularity of the \textit{Cloud Systems}, \textit{GAs} can be executed in parallel without changing their main principle. Currently, one available distributed platform is \textit{Apache Hadoop} and its easy installation and maintainability are two key aspects that contributed to its great popularity. Nowadays it is common for an industry to rent a cluster on-line in order to request the execution of their applications as services.
	
	All these elements add to the motivations described in this paper. \textit{elephant56}\footnote{The name \enquote{elephant56} combines two ideas: \enquote{elephant} resembles the \textit{Hadoop} platform; \enquote{56} is the number of chromosomes of elephants citing the world of \textit{Genetics}.} is a framework for developing \textit{GAs} that can be executed on the \textit{Hadoop} platform, following the paradigm of \textit{MapReduce}. The main purpose of the framework is to completely hide the inner workings of \textit{Hadoop} and allow users to focus on the main aspects of \textit{GA} of their applications, being sure that their task will be correctly executed and with a good performance. In this way, the only concerns of users are the \textit{GA} model and the settings of the key inputs, parameters and functions used in \textit{GAs} (such as \textit{fitness}, \textit{selection}, initial population, etc...).
	
	The intended goals of this paper are:
	\begin{itemize}
		\item Develop a complete framework that allows the user to develop and execute full applications;
		\item Provide some frequent ready \enquote{on-the-shelf} functions, such as common criteria of \textit{selection} and \textit{individuals} representations that can be useful in most of the possible \textit{GAs} implementations;
		\item Develop testing strategies to evaluate the performance of \textit{elephant56}. For this reason a complete example of use of the framework is included in this paper.
	\end{itemize}
	
	The rest of paper is organised as follows. Section \ref{sec:Background and Related Work} shows the basis needed to understand the contents of this paper, with a quick overview of the existing literature about the subject. In section \ref{sec:Design and Development} the design and development of the framework are explained, by starting with the \textit{Driver} component and finishing with the description of all components under two main profiles: the \enquote{core} and \enquote{user}. In section \ref{sec:Preliminary Analysis} the framework is tested by developing a complete example of use called \enquote{Feature Selection Subset}, by explaining how the problem was adapted and showing the results and performance. Section \ref{sec:Conclusions and Future Work} gives a final view of the achieved results suggesting possible future work to improve the framework.

%% file: background_and_related_work.tex
\section{Background and Related Work}
\label{sec:Background and Related Work}

	In the following it is briefly introduced the background of the theory and technologies behind the framework. Furthermore, it is described the developments and ideas of the existing literature in the same area of research.
	
	\subsection{Genetic Algorithms}
		
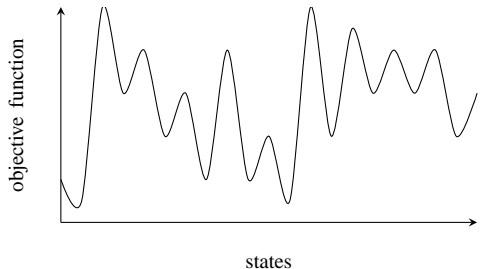
\begin{figure}[htbp]
			\centering
			\input{figures/background_and_related_work/noisy_state-space_landscape}
			\caption{A noisy state-space landscape}
			\label{fig:A noisy state-space landscape}
		\end{figure}

		As mentioned above \textit{Genetic Algorithms}, described in \cite{Goldberg1989} as a metaheuristic technique, can find reasonable solutions within reasonable time, for which exact techniques are unsuitable. \textit{GAs} simulate several aspects of the \enquote{Darwin's Evolution Theory} in order to enable them to converge towards optimal or near-optimal solutions. The main concept behind these metaheuristic algorithms is \textit{robustness}, the balance between efficiency and efficacy necessary for survival in many different environments. Unfortunately, in real-world applications the search is fraught with discontinuities and vast multimodal, in such a way that the space of search is noisy as shown in figure \ref{fig:A noisy state-space landscape}. In order to move carefully through the \textit{space of search}, \textit{heuristics} are needed. Some widely accepted search procedures simply lack of this quality but this does not imply that these algorithms are not useful. Indeed, they have been used successfully in many applications but where the domain of search is limited.
					
		\textit{GAs} differ from more traditional metaheuristic techniques in many ways. They:
		\begin{itemize}
			\item Work with a coding of the parameter set, not the parameters themselves;
			\item Search from a population of states, not a single point;
			\item Use the objective function (\textit{fitness function}) information, not auxiliary knowledge;
			\item Use probabilistic transition rules, not deterministic rules.
		\end{itemize}

		In \textit{GAs} the new individuals are generated by two parents selected among the current population individuals for so-called \textit{sexual reproduction}.
		
		\begin{figure}[htbp]
			\centering
			\includegraphics[width=\linewidth]{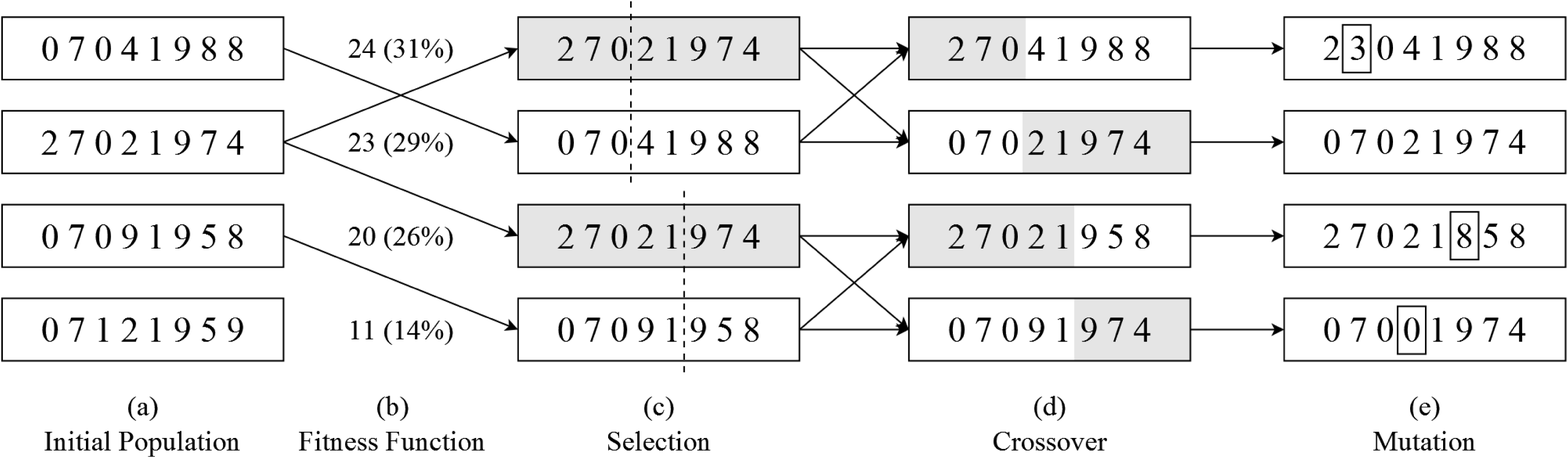}
			\caption[The execution of a Genetic Algorithm]{The execution of a \textit{Genetic Algorithm} (Adapted from \cite{Russell2010})}
			\label{fig:The execution of a Genetic Algorithm}
		\end{figure}
		
		As it can be seen in figure (\ref{fig:The execution of a Genetic Algorithm} (a)), \textit{GAs} begin with a set of $k$ randomly generated states called \enquote{population}. Each state is called \enquote{individual} and is represented as a string over a finite alphabet. This string is called \enquote{chromosome} and each symbol \enquote{gene}. Every iteration of the algorithm generates a new population and consists of the following steps:
		\begin{itemize}
			\item \textbf{Selection:}
			Each state is evaluated by the \textit{fitness function} (b). Each value influences the random choice among the successors for the next step. Once chosen the $k$ successors are grouped into couples (c);
			
			\item \textbf{Crossover:}
			The algorithm chooses for each individual a division point, which is called \textit{crossover point}. At this point the sexual reproduction (d) in which two children are created begins: the first takes the first part of the first parent and the second of the second parent; the other takes the second part of the first  parent and first part of the second parent;
			
			\item \textbf{Mutation:}
			When the \textit{offsprings} are generated, each \textit{gene} is subjected to a random \textit{mutation} with a small independent probability (e).
		\end{itemize}
		
		Depending of the code used to represent the feasible solutions of a given problem, the representation of the genetic code can condition the \textit{crossover} and \textit{mutation} steps. For example, if the \textit{chromosome} is composed of concatenated values, a simple division may not make any sense. The mathematical explanation of why \textit{GAs} work, was given by the \enquote{Holland's Schema Theorem} in \cite{Holland1995}, which says:
		\begin{quote}
			\itshape
			Short, low order, above average schemata receive exponentially increasing trials in subsequent generations of a Genetic Algorithm.
		\end{quote}
			
		The concept of \textit{schema} is a string (\textit{chromosome}) in which some of the positions (\textit{genes}) can be left unspecified. For example, with a binary alphabet representation, the \textit{schema} $01*0$ describes all \textit{chromosomes} (\textit{instances} of the \textit{schema}) in which the first, the third and the fourth \textit{genes} are fixed, which are $0100$ and $0110$. \textit{Holland} showed that if the average \textit{fitness value} of the instances of a \textit{schema} is above the average, the number of instances of the \textit{schema} within the population will grow over time. This justifies the choice of selecting with more probability individuals that have higher \textit{fitness value} and the need to use the \textit{mutation} to shake things up. Moreover, it is shown that since each individual owns a large number of different \textit{schemas}, during each generation the number of \textit{schemas} implicitly processed are in the order of $k^3$, where $k$ is the number of individuals. Understanding if \textit{GAs} provide near-optimal solutions is the object of study of the \enquote{Building-Block Hypothesis} which has to be confirmed yet.
			
	\subsection{Hadoop MapReduce}
		
		The term \textit{Hadoop} \cite{White2012} comprises a family of many related projects with the same infrastructure for distributed computing and large-scale data processing. It is better known for the \textit{MapReduce} algorithm, shown below, and its distributed file system \textit{HDFS}, which runs on large clusters of commodity machines. \textit{Hadoop} was created by \textit{Doug Cutting} and has its origins in \textit{Apache Nuts}, an open source web search engine. In January 2008 \textit{Hadoop} was made a top-level project at \textit{Apache}, attracting to itself a large active community, including \textit{Yahoo!}, \textit{Facebook} and \textit{The New York Times}. At present, \textit{Hadoop} is a solid and valid presence in the world of cloud computing.
		
		\textit{MapReduce} is a programming model whose origins lie in the old functional programming. It was adapted by \textit{Google} \cite{Dean2004} as a system for building search indexes, distributed computing and database communities. It was written in \textit{C++} language and was made as a framework, in order to simply develop its applications. In \textit{Hadoop} programs are mainly in \textit{Java} language but it is also possible, through a mechanism called \enquote{streaming}, to develop programs in any language that supports the \textit{standard I/O}. \textit{MapReduce} is a batch query processor and the entire dataset is processed for each query. It is a linearly scalable programming model where users programs at least two functions: the \enquote{map} function and \enquote{reduction} functions. These functions process the data in terms of key/value pairs which are unaware of the size of the data or the cloud that they are operating on, so they can be used unchanged either for a small dataset or for a massive one.
			
		\begin{figure}[htbp]
			\centering
			\includegraphics[width=\linewidth]{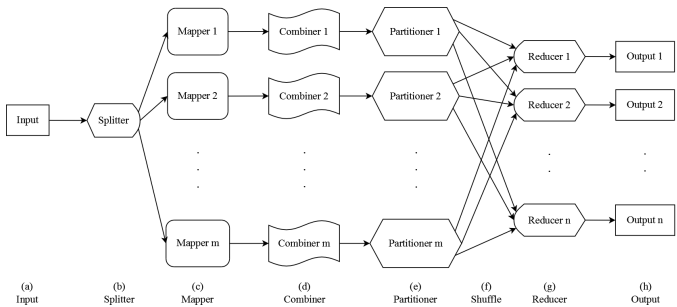}
			\caption{\textit{Hadoop MapReduce}}
			\label{fig:Hadoop MapReduce}
		\end{figure}
			
		A typical \textit{MapReduce} program (figure \ref{fig:Hadoop MapReduce}) starts on a single machine using a piece of code called \enquote{driver}. This launches and manages the execution called \enquote{job} of the entire distributed program on the cluster. Then several components at different stages operate:
		\begin{itemize}
			\item \textbf{Splitter:}
			The input (a) is often in the form of a simple text, which consists of one or more files stored in the distributed file system \textit{HDFS}. It undergoes a first treatment from the \textit{Splitter} (b). Depending on the criteria used, it first creates the key/value pairs called \enquote{records} and sends them directly to an available \textit{Mapper} on the cluster. The function, where $k_1$ and $v_1$ means data types, is described as:
			\begin{equation*}
				\text{split} : \textit{input} \to \operatorname{list}\left(\textit{k}_{1}, \textit{v}_{1}\right)_{S}
			\end{equation*}
			
			\item \textbf{Mapper:}
			It is the first function of the core of \textit{Hadoop}. A \textit{Mapper} (c) runs on a machine of the cluster a process called \enquote{MapTask}. Once the input has been given, it produces a list of records according to the algorithm described by the programmer:
			\begin{equation*}
				\text{map} : \left(\textit{k}_{1}, \textit{v}_{1}\right)_{S} \to \operatorname{list}\left(\textit{k}_{2}, \textit{v}_{2}\right)_{M}
			\end{equation*}
			
			\item \textbf{Combiner:}
			It is also called \enquote{Local Reducer} and it is an optional component. The \textit{Combiner} function does not replace the \textit{reduce} function but it can help cutting down the amount of data exchanged between \textit{Mappers} and \textit{Reducers}. It (d) runs on the same machine that made the \textit{MapTask} and it computes new pairs having the same key:
			\begin{equation*}
				\text{combiner} : \left(\textit{k}_{2}, \operatorname{list}\left(\textit{v}_{2}\right)\right)_{M} \to \left(\textit{k}_{2},\textit{v}_{2}\right)_{M^1}
			\end{equation*}
			
			\item \textbf{Partitioner:}
			It (e) establishes the criteria by which records are assigned to a \textit{Reducer}. This is also called the \enquote{shuffle operation} (f) and ensures that records with the same key will be assigned to the same \textit{Reducer}. If not directly specified, the default \textit{Partitioner} acts like a hash function on keys:
			\begin{equation*}
				\text{partitioner} : \textit{k}_{2} \to \operatorname{hash}\left(\textit{k}_{2}\right)
			\end{equation*}
			
			\item \textbf{Reducer:}
			Finally, the \textit{Reducer} (g) concludes the job. If a \textit{Partitioner} with hash on keys has been used, it can process all the records with the same key created by the whole cluster:
			\begin{equation*}
				\text{reduce} : \left(\textit{k}_{2}, \operatorname{list}\left(\textit{v}_{2}\right)\right)_{M^1} \to \left(\textit{k}_{3}, \textit{v}_{3}\right)_R
			\end{equation*}
		\end{itemize}
		
		Once the job has been completed, the last records are written one for each \textit{Reducer} to the output files in \textit{HDFS} (h).

	\subsection{Related Work}
	
		The existent literature proposes some parallel version of \textit{GAs} using the \textit{MapReduce} paradigm. The first is an extension, by adding a second \textit{Reducer}, of \textit{MapReduce} named \enquote{MRPGA} \cite{Jin2008} based on \textit{.Net}. In this implementation a coordinator client manages the executions of the parallel \textit{GA} iterations. The chosen model is the island model in which each participating node computes \textit{GAs} operations for a portion of the entire population. In the first phase, each \textit{Mapper} node receives its own portion of population and computes the \textit{fitness value} for each of its individuals. The \textit{Reducer} nodes of the first reduce phase receive the individuals of the correspondent island and apply the \textit{selection} function. The final \textit{Reducer} computes the global \textit{selection} and the other following \textit{GAs} functions.
			
		Another approach, this time developed on \textit{Hadoop}, is presented by \cite{Verma2009}. The number of \textit{Mapper} nodes and the one of the \textit{Reducer} nodes are unrelated. The \textit{Mapper} nodes computes the \textit{fitness} function and the \textit{Reducer} a local \textit{selection} followed by the other \textit{GAs} functions. The substantial difference with \textit{MRPGA} lies in the fact that the partitioner supplies a sort of \enquote{migration} among the individuals as it randomly sends the outcome of the \textit{Mapper} nodes to different \textit{Reducer} nodes.
		
		In \cite{DiMartino2012} the three main grain parallelism are described and implemented by exploiting the \textit{MapReduce} paradigm:
		\begin{enumerate}
			\item \textit{Fitness Evaluation Level} (\textit{Global Parallelisation Model});
			\item \textit{Population Level} (\textit{Coarse-grained Parallelisation Model} or \textit{Island Model});
			\item \textit{Individual Level} (\textit{Fine-grain Parallelisation Model} or \textit{Grid Model}).
		\end{enumerate}
		
\begin{figure}[htbp]
			\centering
			\includegraphics[width=0.8\linewidth]{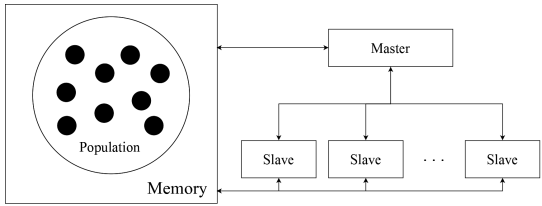}
			\caption[Global Parallelisation Model]{\textit{Global Parallelisation Model} (Adapted from \cite{Luque2011})}
			\label{fig:Global Parallelisation Model}
		\end{figure}

		In the \textit{Global Parallelisation Model} (figure \ref{fig:Global Parallelisation Model}) a master node manages the population and computes for them all the \textit{GAs} functions but the fitness evaluation is computed by the slave nodes. The model is adapted to \textit{MapReduce} by delegating some \textit{Mappers} the task of evaluating the \textit{fitness value} for each individual in parallel. Then, the single \textit{Reducer} collects the results and performs the other \textit{GA} operations. One generation corresponds to one \textit{MapReduce} execution, so that the whole computation is a sequence of \textit{MapReduce} executions.
		
\begin{figure}[htbp]
			\centering
			\includegraphics[width=0.9\linewidth]{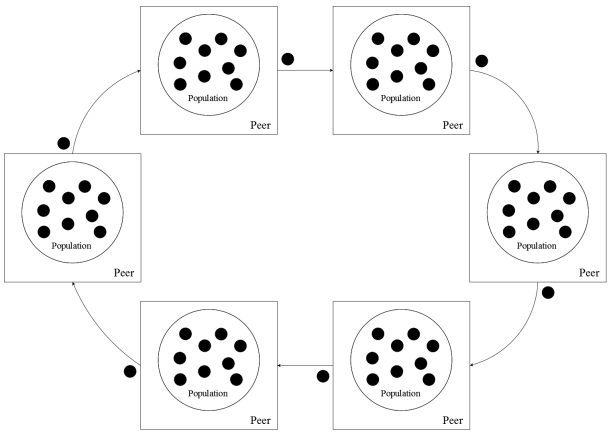}
			\caption[Coarse-grain Parallelisation Model]{\textit{Coarse-grain Parallelisation Model} (Adapted from \cite{Luque2011})}
			\label{fig:Coarse-grain Parallelisation Model}
		\end{figure}

		In the \textit{Coarse-grained Parallelisation Model} (figure \ref{fig:Coarse-grain Parallelisation Model}) the population is subdivided into \enquote{islands} and the \textit{GA} is independently run on each of them. Periodically the islands exchange information by \enquote{migrating} some individuals. Here the number of \textit{Reducers} is higher than in the previous model. After having computed the \textit{fitness values} in the \textit{Mappers}, a \textit{Partitioner} assigns each island to a different \textit{Reducer} in order to compute the other \textit{GAs} functions in parallel.
		
\begin{figure}[htbp]
			\centering
			\includegraphics[width=0.7\linewidth]{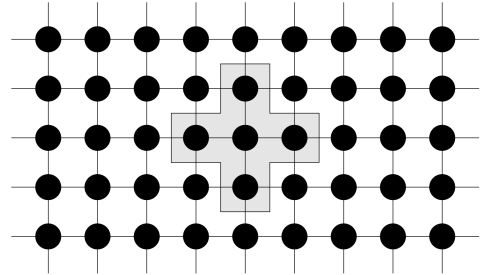}
			\caption[Fine-grain Parallelisation Model]{\textit{Fine-grain Parallelisation Model} (Adapted from \cite{Luque2011})}
			\label{fig:Fine-grain Parallelisation Model}
		\end{figure}

		In the \textit{Fine-grain Parallelisation Model} (figure \ref{fig:Fine-grain Parallelisation Model}) each individual is placed on a grid and the \textit{GA} operations are performed in parallel by evaluating simultaneously the \textit{fitness value} and applying the \textit{selection} limited only to the small adjacent neighbourhood. This model is slightly adapted by modifying the previous one: the \textit{Partitioner} uses a pseudo-random function in such a way that the described local neighbourhood is developed.
	
		The \textit{Global Parallelisation Model} has been implemented by \cite{DiMartino2012} in order to solve a problem of \textit{Automatic Test Data Generation}. It has been developed on the \textit{Google App Engine MapReduce} platform. Also serving the same purpose, \cite{DiGeronimo2012} has developed \textit{Coarse-grained Parallelisation Model} on the \textit{Hadoop MapReduce} platform.
		
		The framework presented in this paper mainly exploits the \textit{Coarse-grained Parallelisation Model}, without differing so much from the implementation proposed by \cite{DiMartino2012}. It introduces some new modifies that better suit its intrinsic nature of framework.

%% file: figures/background_and_related_work/noisy_state-space_landscape.tex
\begin{tikzpicture}
	\begin{axis}
		[
			width=0.8\linewidth,
			height=0.5\linewidth,
			axis x line=bottom,
			axis y line=left,
			xlabel={\footnotesize states},
			ylabel={\footnotesize objective function},
			xticklabel=\empty,
			yticklabel=\empty,
			xmin=0,
			ymin=0,
			xmax=20,
			ymax=5,
			tick style={draw=none}
		]
		\addplot
			[domain=2:10, samples=200, smooth]
			coordinates
			{(0, 1) (1, 0.5) (2, 5) (3, 3) (4, 4) (5, 2) (6, 3) (7, 1) (8, 4) (9, 1) (10, 2) (11, 0.5) (12, 5) (13, 2) (14, 4.5) (15, 3) (16, 4) (17, 3) (18, 4) (19, 2) (20, 3)};
	\end{axis}
\end{tikzpicture}

%% file: design_and_development.tex
\section{Design and Development}
\label{sec:Design and Development}

	Here the design and development of the single involved components are explained. It starts from the last level of abstraction, which is the \textit{Driver} component, and goes into depth as far as the \textit{Hadoop} development of the components.
	
	\subsection{Driver}
	
		The \textit{Driver} is the main part of the framework and it represents the only interface between the two involved parts in each work, managing the two main aspects:
		\begin{itemize}
			\item The interaction with the user;
			\item The launch of the jobs on the cluster. 
		\end{itemize}
		
		The \textit{Driver} is executed on a machine even separated from the cluster and it also computes some functions as soon as the first results are received from the cluster.
	
		\begin{figure}[htbp]
			\centering
			\includegraphics[width=0.75\linewidth]{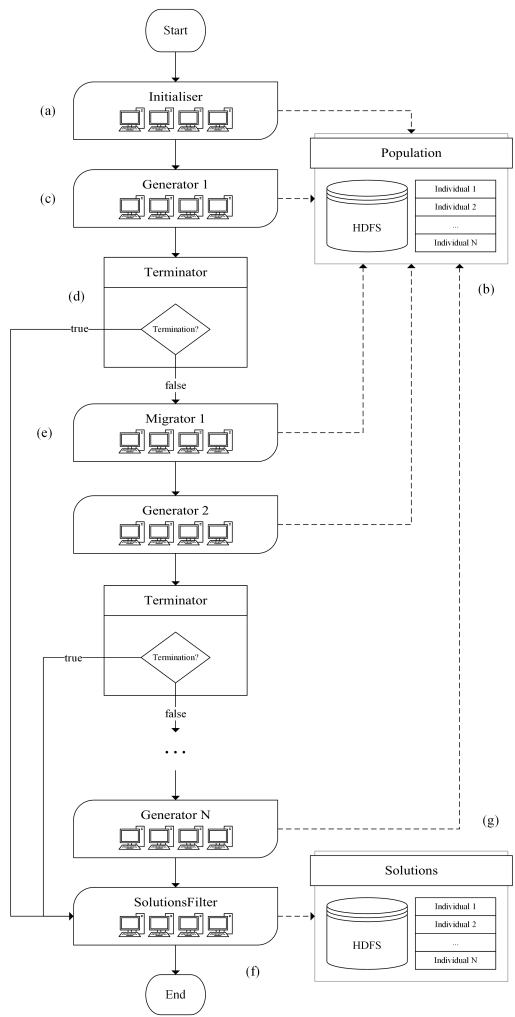}
			\caption{The \textit{Driver}}
			\label{fig:The Driver}
		\end{figure}
		
		Figure \ref{fig:The Driver} describes the elements of the whole process every time it is executed, each one managed by the \textit{Driver} process:
		\begin{itemize}
			\item \textbf{Initialiser:}
			This component (a) can be optional if there already exists a population. The user can define how to generate the first individuals for each island. The default component makes it randomly. The \textit{Initialiser} is a \textit{MapReduce} job which produces the individuals in the form of a sequence of files, stored in a serialised form directly in \textit{HDFS} (b);
			
			\item \textbf{Generator:}
			It is the heart of the framework (c). It executes one generation on the cluster and produces the new population, storing each individual again in \textit{HDFS} within the fitness values and a flag indicating if an individual satisfies the \textit{termination criterion}.
			
			\item \textbf{Terminator:}
			At the end of each generation, the \textit{Terminator} component (d), which does not work as a job, checks the stopping conditions (e.g., if the maximum number of generations has been reached). Once terminated, the population is directly submitted to the \textit{SolutionsFilter} job (f).
			
			\item \textbf{Migrator:}
			This optional job (e) allows moving individuals from an island to another, according to the criteria defined by the user such as the frequency of migration, number and destinations of migrants and the selection method for choosing migrants.
			
			\item \textbf{SolutionsFilter:}
			When the job terminates, all the individuals of the last generation are filtered according to those that satisfy the \textit{termination criterion} and those which do not. Then, the results of the whole process is stored in \textit{HDFS} (g).
		\end{itemize}
				
	\subsection{Generator}
			
		Each \textit{generator} job makes the population evolve. In order to develop the complex structure described below, it is needed to use multiple \textit{MapTasks} and \textit{ReduceTasks}. This is possible using a particular version of \textit{ChainMapper} and \textit{ChainReducer} classes of Hadoop, slightly modified in order to treat \textit{Avro} objects. Hadoop manages the exchange of information among the tasks with a raw method of serialisation. Using \textit{Avro} \cite{Cutting2011}, it is easy to store objects and to save space on the disk, allowing an external treatment of them. It also permits a quick exchange data among the parties involved in the \textit{MapReduce} communication.
					
		A chain allows to manage the tasks in the form described by the pattern:
		\begin{equation*}
			\left(\textit{MAP}\right)^+\left(\textit{REDUCE}\right)\left(\textit{MAP}\right)^*
		\end{equation*}
		which means one or more \textit{MapTasks}, followed by one \textit{ReduceTask} and other possible \textit{MapTasks}.

		\begin{figure}[htbp]
			\centering
			\includegraphics[width=0.7\linewidth]{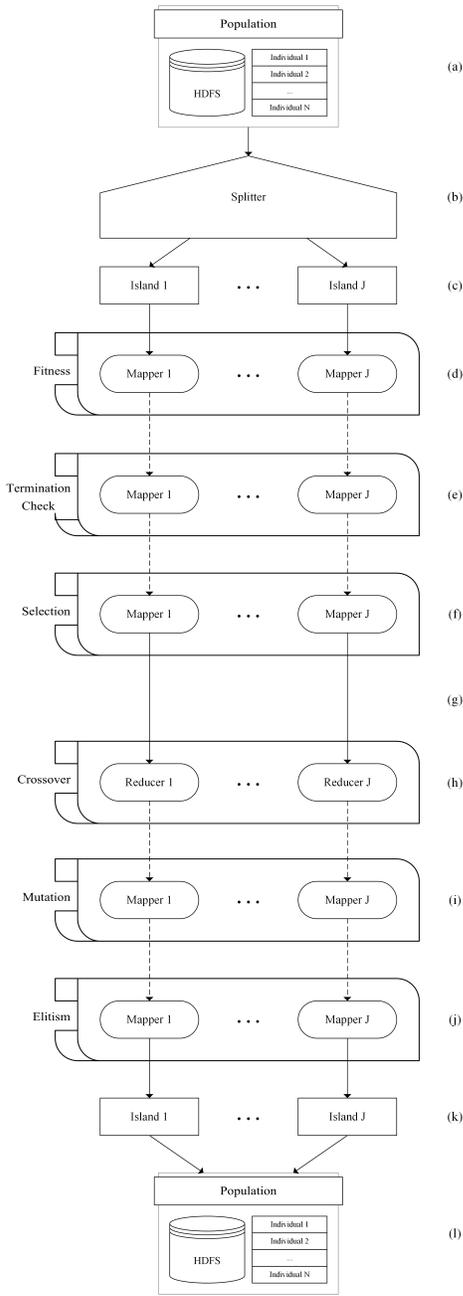}
			\caption{The \textit{Generator} component}
			\label{fig:The Generator component}
		\end{figure}
		
		Figure \ref{fig:The Generator component} describes how the generation work is distributed on the cloud:
		\begin{itemize}
			\item \textbf{Splitter:}
			The \textit{Splitter} (b) takes as input the \textit{population} (a), deserialises it and splits the individuals into $J$ groups (islands) (c), according to the order of individuals. Each split contains a list of records, one per each individual:
			\begin{equation*}
				\text{split} : \textit{individuals} \to \operatorname{list}\left(\textit{individual}, \textsc{null}\right)
			\end{equation*}
			
			During the deserialisation, the splitter adds some fields to the objects which will be useful for the next steps of the computation, such as the \textit{fitness} function.
			
			\item \textbf{Fitness:}
			Here (d) according to $J$ islands, the $J$ \textit{Mappers} compute the \textit{fitness values} for each individual within its corresponding island for which it has not been calculated yet:
			\begin{equation*}
				\text{map} : \left(\textit{individual}, \textsc{null}\right) \to \left(\textit{individual}_{\text{F}}, \textsc{null}\right)
			\end{equation*}
			
			The user defines how to the \textit{fitness} is evaluated and the values are stored inside the corresponding field of the objects.
			
			\item \textbf{TerminationCheck:}
			Without leaving the same machine of the previous \textit{map}, the second \textit{map} (e) acts in a chain. It checks if the current individuals satisfy the \textit{termination criterion}. This is useful, for instance, when a lower limit target of the \textit{fitness value} is known. If at least one individual gives a positive answer to the test, the event is notified to the other phases and islands by a flag stored in \textit{HDFS}. This avoids the executions of the next phases and generations.
			
			\item \textbf{Selection:}
			If the \textit{termination criterion} has not been satisfied yet, this (f) is the moment to choose the individuals that will be the parents during the \textit{crossover} for the next iteration. The users can define this phase in their own algorithms. The couples which have been selected are all stored in the key:
			\begin{equation*}
				\begin{gathered}
					\text{map} : \left(\textit{individual}_{\text{F}}, \textsc{null}\right) \to \\
					\left(\textit{couple\_information}, \textit{individual}_{\text{F}}\right)
				\end{gathered}
			\end{equation*}
			
			If an individual has been chosen more than one time, it is replicated. Then all the individuals, including those not chosen if the \textit{elitism} is active, leave the current worker and go to the correspondent \textit{Reducer} for the next step (g).
			
			\item \textbf{Crossover:}
			In this phase (h), the individuals are grouped by the couples established during the \textit{selection}. Then each \textit{Reducer} applies the criteria defined by the user and makes the \textit{crossover}:
			\begin{equation*}
				\begin{gathered}
					\text{reduce} : \left(\textit{couple\_information}, \operatorname{list}\left(\textit{individual}_{\text{F}}\right)\right) \to \\
					\left(\textit{individual}, \textsc{true}\right)
				\end{gathered}
			\end{equation*}
			
			This produces the \textit{offspring} (marked with the value \textsc{true} in the value field) that is read, together with the previous population, during the next step.
			
			\item \textbf{Mutation:}
			During the \textit{mutation} (i), the chained \textit{Mappers} manage to make the mutation of the genes defined by the user. Only the \textit{offspring} can be mutated:
			\begin{equation*}
				\text{map} : \left(\textit{individual}, \textsc{true}\right) \to \left(\textit{individual}_{\text{M}}, \textsc{null}\right)
			\end{equation*}
			
			\item \textbf{Elitism:}
			In the last phase (j), if the user chooses to use elitism, the definitive population is chosen among the individuals of the offspring and the previous population:
			\begin{equation*}
				\text{map} : \left(\textit{individual}, \textsc{null}\right) \to \left(\textit{individual}, \textsc{null}\right)
			\end{equation*}
			
			At this point (k), the islands are ready to be written in \textit{HDFS} (l).
			
		\end{itemize}
		
\begin{figure*}[htbp]
			\centering
			\includegraphics[width=0.8\textwidth]{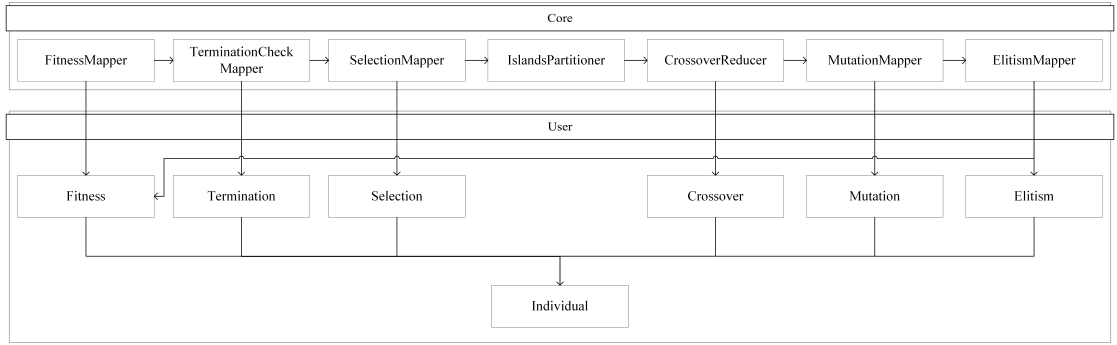}
			\caption{The two levels of abstraction of the \textit{Generator} component}
			\label{fig:The two levels of abstraction of the Generator component}
		\end{figure*}

		The architecture of the generator component provides two levels of abstraction (see figure \ref{fig:The two levels of abstraction of the Generator component}):
		\begin{itemize}
			\item The first one, which is called the \enquote{core} level, allows the whole job to work;
			\item The second one, which is called the \enquote{user} level, allows the user to develop his own \textit{Genetic Algorithm} and to execute it on the cloud.
		\end{itemize}
	
		The \textit{core} is the base of the framework with which the end-user does not need to interact. Indeed, the fact that a \textit{MapReduce} job is executed is totally invisible to the user. It consists of everything that is needed to run an application on \textit{Hadoop}. The final user can develop his own \textit{GA} simply by implementing the relative classes, without having to deal with \textit{map} or \textit{reduce} details. If the user does not extend these classes, a simple behaviour is implemented by doing nothing else than forwarding the input as output. The framework also makes some default classes available so that the user can use them for most cases.
				
	\subsection{Terminator}
	
\begin{figure}[htbp]
			\centering
			\includegraphics[width=0.6\linewidth]{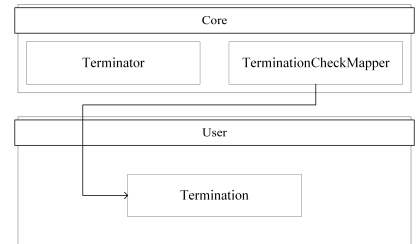}
			\caption{The two levels of abstraction of the \textit{Terminator} component}
			\label{fig:The two levels of abstraction of the Terminator component}
		\end{figure}

		The \textit{Terminator} component (figure \ref{fig:The two levels of abstraction of the Terminator component}) plays two roles:
		\begin{itemize}
			\item After the execution of every generation job, by calling the methods of the \textit{Terminator} class on the same machine where the \textit{Driver} is running;
			\item During the generator job, through the use of the \textit{TerminationCheckMapper}.
		\end{itemize}
		
		It checks if the stopping conditions have occurred:
		\begin{itemize}
			\item the count of the maximum number of generations has been reached;
			\item at least one individual has been marked of satisfying the termination criterion during the most recent generation phase.
		\end{itemize}
		
		The count is maintained by storing a local counter variable that is updated after the end of each generation. The check for the presence of marked individuals is done by looking for possible flags in \textit{HDFS}. If it terminates, the execution of the \textit{SolutionsFilter} component will eventually follow.
		
	\subsection{Initialiser}
	
\begin{figure}[htbp]
			\centering
			\includegraphics[width=0.8\linewidth]{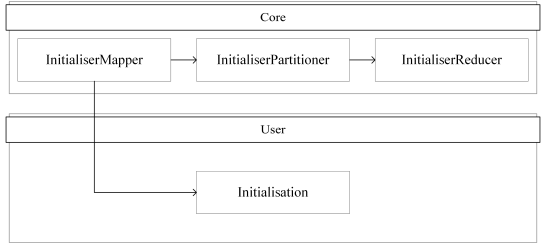}
			\caption{The two levels of abstraction of the \textit{Initialiser} component}
			\label{fig:The two levels of abstraction of the Initialiser component}
		\end{figure}

		The \textit{Initialiser} (see figure \ref{fig:The two levels of abstraction of the Initialiser component}) computes an initial population. This is an optional component, because the entire work can start even if the population data are already present inside \textit{HDFS}.
		
		The class \textit{InitialiserMapper} generates the next individual according to the user's definition.
		
	\subsection{Migrator}
	
\begin{figure}[htbp]
			\centering
			\includegraphics[width=0.8\linewidth]{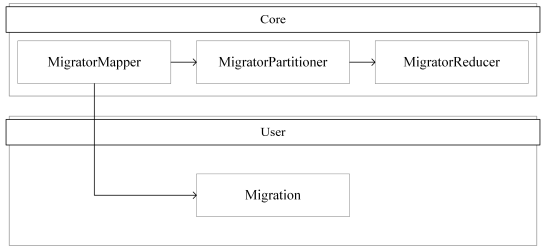}
			\caption{The two levels of abstraction of the \textit{Migrator} component}
			\label{fig:The two levels of abstraction of the Migrator component}
		\end{figure}

		The optional \textit{Migrator} component (figure \ref{fig:The two levels of abstraction of the Migrator component}) shuffles individuals among the islands according to the user's definition. It is at the same time a local component and a job executed on the cloud.
	
		It is started by the \textit{Driver} and on the frequency counter, chooses if it is time to perform a migration.
			
	\subsection{SolutionsFilter}
	
\begin{figure}[htbp]
			\centering
			\includegraphics[width=0.8\linewidth]{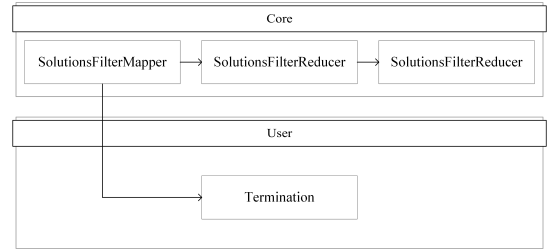}
			\caption{The two levels of abstraction of the \textit{SolutionsFilter} component}
			\label{fig:The two levels of abstraction of the SolutionsFilter component}
		\end{figure}

		The component \textit{SolutionsFilter} (figure \ref{fig:The two levels of abstraction of the SolutionsFilter component}) is invoked only when at least one individual has provoked the termination by satisfying the \textit{termination criterion}. It simply filters the individuals of the last population by dividing those that satisfy the criterion from those which do not.
		
%% file: preliminary_analysis.tex
\section{Preliminary Analysis}
\label{sec:Preliminary Analysis}

	In order to test the performance of the framework, a real-world problem has been used, recording results and performance. The problem of classification in \textit{Machine Learning} consists of learning from well-known example data, called \enquote{training dataset}, with the purpose of being able to properly classify any new input data. The existing algorithms of \textit{Machine Learning} act like predictors on the new data. Therefore it is important to have a good level of \enquote{accuracy} of the prediction. One way of improving its performance is to find an optimal dataset, resulting from the original \textit{training dataset}. This problem is known as \enquote{Feature Subset Selection}. Unfortunately, the search for the optimum is a \textit{NP-Hard} problem. \textit{Genetic Algorithms} have been used to model and look for near-optimal classes for the problem.

	\subsection{Feature Subset Selection}
	
		The \textit{training dataset} includes a list of records, also called \enquote{instances}. Each record is a list of \enquote{attributes} (\textit{features}), in which one attribute is the \textit{class attribute}. Given a \textit{training dataset}, the next step is to build a classifier. For instance, the \textit{C4.5} algorithm can be used in order to build a \textit{Decision Tree}, which is able to give a likely class of ownership for every record in the new dataset. The effectiveness of the classifier is measured by the \textit{accuracy} of the classification of the new data:
		\begin{equation*}
			\textit{accuracy} = \frac{\text{correct classifications}}{\text{total of classifications}}
		\end{equation*}
		
		Sometimes, it is possible to have the same \textit{accuracy} or even a better one with a classifier that considers a lower number of attributes rather than one with more, as explained in \cite{Hall1999}.
		
		The search for an optimal \textit{feature subset} is not only important for building the classifier quickly and with very good response time, but also in terms of real costs. For example, let the problem to be solved be the identification of the presence of a certain disease. The \textit{training dataset} is a collection of real medical records, where some features are cost-free information about the patient, whereas others are expensive test results such as blood test, \textit{DNA} test, etc. In such a context, it is clear that saving the number of collected features means saving money.
		
		Once outlined the advantages of reducing the number of features, describing the problem under a mathematical point of view is needed. Having a \textit{training dataset} with $m$ different attributes, except for the \textit{class attribute} which has to be always present, the number of possible subsets is $2^{m}$. Let $C$ be the execution time in order to build a classifier, which has received a dataset, and let $A$ be the execution time to compute the \textit{accuracy} of the resulting classifier, the resulting running time to find the best subset would be:
		\begin{equation*}
			O\left(2^{m}\right)AC
		\end{equation*}
		As a consequence, the use of a \textit{GA} may simplify the search within the space of solutions and give the proper way to look for good near-optimal solutions.
		
		The resulting model is an adaptation of the application will be executed by the framework where each part is modelled based on the \textit{Feature Subset Selection}. The \textit{driver} is the main part of the algorithm and is executed on one machine. Moreover, who is going to use the algorithm must specify for the execution the following information:
		\begin{itemize}
			\item The generic arguments for the \textit{GA} execution, such as how many individuals initially to generate, the maximum number of generations, etc.;
			\item The \textit{training dataset};
			\item The \textit{test dataset}.
		\end{itemize}
		
		There is also the possibility of specifying the \textit{accuracy} target in order to terminate the process before reaching the maximum number of generations, if this lower limit has been reached during the computation.	
		
		Since computing the \textit{accuracy} for all the possible subsets needs exponential time, the idea is to submit a randomly generated initial group of attribute subsets, making the initial population, to the algorithm generations. During each generation, every subset is evaluated by computing the \textit{accuracy} value and all the \textit{GAs} functions are applied until target \textit{accuracy} is achieved or maximum number of generations is reached. The control of the satisfying the \textit{termination criterion} is controlled during the \textit{termination} phase. At the end, the last population is ready to be tested with the \textit{test dataset}.
	
		By giving the \textit{training dataset} as input, the first applied operation is the \textit{initialisation}. The algorithm generates the $r$ random attribute subsets which will be the initial population of individuals. Every individual (subset) is encoded as an array of $m$ bit, where each bit shows if the corresponding enumerated attribute is present into the subset (value $1$) or not (value $0$). Since the records in the \textit{training dataset} are never altered during the whole algorithm, it is not necessary to encode them. They will be available when needed for the generations.
		
		Every generation phase processes all the individuals within the population, according to the following steps:
		\begin{itemize}
			\item \textbf{Fitness:}
			For each subset of attributes, the \textit{training dataset} is filtered to respect the attributes in the subset. In such a way, the fitness value is computed by applying the steps:
			\begin{enumerate}
				\item Select the current portion of the dataset that is going to act as \textit{training dataset} and the portion as \textit{test dataset};
				\item Build the \textit{Decision Tree} through the \textit{C4.5} algorithm and computing the \textit{accuracy} by submitting the current dataset;
				\item The operations are repeated according to the \textit{folding parameter}, following the technique of the \enquote{Crossing Folding};
				\item The best \textit{accuracy} is returned.
			\end{enumerate}
			
			\item \textbf{Selection:}
			It chooses which individuals will be the parents during the \textit{crossover}. It is important to give the individuals with the best \textit{accuracy} the best probability to be chosen. The algorithm uses the method of the \enquote{roulette-wheel selection}: it builds a wheel according to the \textit{fitness values} (\textit{accuracy}) of each individual, after which it will turn for every new couple in such a way as to choose who will form it.
			
			\item \textbf{Crossover:}
			At this step, the new \textit{offspring} is produced splitting the parents of each couple into two parts, according to a random \textit{crossover point}, and then mixing the parts obtaining two new children, which have one part of the mother and one of the father.
			
			\item \textbf{Mutation}
			According to a probability to mutate, during this step each subset may change the attributes into itself.
			
			\item \textbf{Elitism}
			Optionally enabled, this step allows to choose the best individuals among the ones in the new \textit{offspring} and in the previous generation, in such a way as to guarantee the growth of the \textit{accuracy} target after every generation.
		\end{itemize}

		Since the algorithm is executed on different islands, it will be important to give a variability factor for each of them. The \textit{Migration} manages the passage of a certain number of randomly selected individuals from an island to another.
	
	\subsection{Subject}
	
		Three example dataset were given, referring to real problems. They are all coming from the \textit{UCI Machine Learning Repository}\footnote{UCI Machine Learning Repository\newline\url{http://archive.ics.uci.edu/ml/}}. Everything was submitted with little differences of parameters in each specific case, but all with two options of \textit{elitism} active and not. Since these experiments aim is to analyse the behaviour of the algorithm during a full computation, the target \textit{accuracy} was not specified to let the algorithm be executed until the maximum number of generations be reached.
	
		Each dataset was divided into two parts:
		\begin{itemize}
			\item The first $60 \%$ as \textit{training dataset};
			\item The last $40 \%$ as \textit{test dataset}, eventually used to compute the obtained best \textit{accuracy}.
		\end{itemize}
		
		The test bench was composed by three versions:
		\begin{itemize}
			\item \textbf{Sequential:}
			A single machine executes a \textit{GA} similar to the framework one by using the \textit{Weka}\footnote{\textit{Weka} is a collection of \textit{Machine Learning} algorithms for data mining tasks. The algorithms can either be applied directly to a dataset or called from \textit{Java} code. \textit{Weka} contains tools for data pre-processing, classification, regression, clustering, association rules, and visualisation. It is also well-suited for developing new \textit{Machine Learning} schemes.\\\url{http://www.cs.waikato.ac.nz/ml/weka/}} \textit{Java} library.
			
			\item \textbf{Pseudo-distributed:}
			Here the framework version of the algorithm make the scene. It is executed on a single machine again, setting the number of islands to one. It requires \textit{Hadoop} in order to be executed.
			
			\item \textbf{Distributed:}
			The framework is executed on a cluster of computers on \textit{Hadoop} platform. This is the version of most interest.
		\end{itemize}
		
		All the versions execute on a remote \textit{Amazon EC2} cluster. This was chosen in order to give a factor of fairness to all the solutions.
		
		The single machine of the \textit{sequential} version was:
		\begin{center}
			\scriptsize
			\input{tables/preliminary_analysis/sequential_version}
		\end{center}
		
		Although it run a \textit{sequential} algorithm, the \textit{pseudo-distributed} version needed a \textit{Hadoop} installation:
		\begin{center}
			\scriptsize
			\input{tables/preliminary_analysis/pseudo-distributed_version}
		\end{center}
		
		On the other hand, the \textit{distributed} version needed a full \textit{Hadoop} cluster:
		\begin{center}
			\scriptsize
			\input{tables/preliminary_analysis/distributed_version}
		\end{center}
			
	\subsection{Results}
		
		It is interesting analysing some aspects from the results. These consist of a measure of the effort in developing the application for \textit{Feature Subset Selection} with the framework and other aspects regarding the performance.
		
		\subsubsection{Developing Effort}
		
\begin{figure}[htbp]
				\centering
				\input{figures/preliminary_analysis/pie_chart_framework_developing}
				\caption{The number of lines of code for the framework developing}
				\label{fig:The number of lines of code for the framework developing}
			\end{figure}
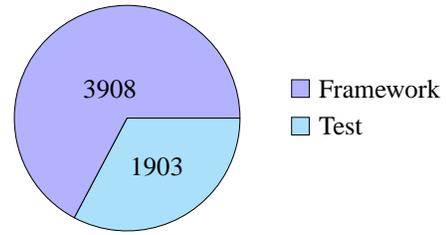

			A total of $5811$ lines of code were written during the development of the whole framework where $1903$ include the test classes, as shown in figure \ref{fig:The number of lines of code for the framework developing}.
			
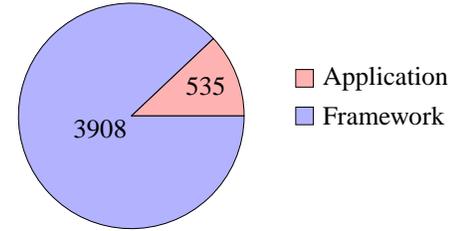
\begin{figure}[htbp]
				\centering
				\input{figures/preliminary_analysis/pie_chart_application_developing}
				\caption{The number of lines of code for the application developing}
				\label{fig:The number of lines of code for the application developing}
			\end{figure}

			For the development of the application (figure \ref{fig:The number of lines of code for the application developing}) of \textit{Feature Subset Selection} were written $535$ lines of code against the $3908$ of the framework infrastructure.
			
		\subsubsection{Number of attributes}
		
			It is the number of attributes of the best individual after the final generation. This parameter is directly referred to the specific problem, because it was one of the target to achieve. As described before, the lower the value is, the better it is.
			
			\begin{figure}[htbp]
				\centering
				\input{figures/preliminary_analysis/plot_stats_attributes}
				\caption{Comparison of the number of attributes}
				\label{fig:Comparison of the number of attributes}
			\end{figure}
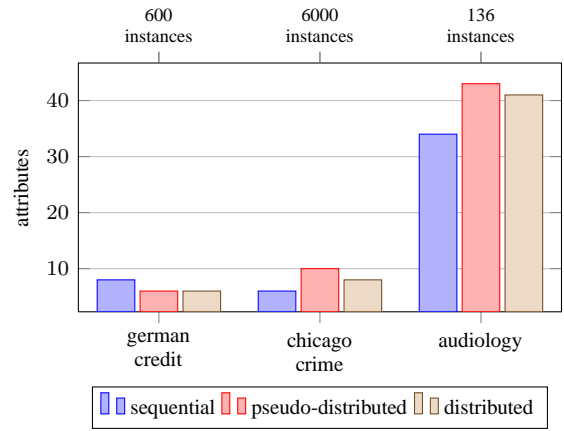
			
			Figure \ref{fig:Comparison of the number of attributes} shows that all the versions act more or less in the same way. It is more important to consider the next parameter.
			
		\subsubsection{Accuracy}
		
			The \textit{accuracy} is the value that was computed just after the execution of each algorithm. It was obtained by submitting a common \textit{test dataset} to the resultant new \textit{training dataset} filtered through the subset of attributes found at the end. 
			
			\begin{figure}[htbp]
				\centering
				\input{figures/preliminary_analysis/plot_stats_accuracy}
				\caption{Comparison of the accuracy}
				\label{fig:Comparison of the accuracy}
			\end{figure}
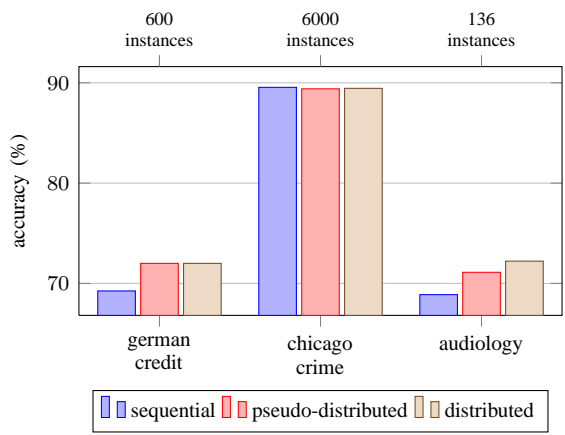
			
			By looking at figure \ref{fig:Comparison of the accuracy}, again the three versions do not have substantial differences. The upside is that the framework achieves its objective, by giving a subset that has both a reduced number of attributes and an \textit{accuracy} that still suits the initial one.
		
		\subsubsection{Running time}
		
			While the number of attributes and \textit{accuracy} give a measure of efficacy of the \textit{distributed} version, the running time weighs against the different versions of the algorithms at the time of choosing which one to use.
			
			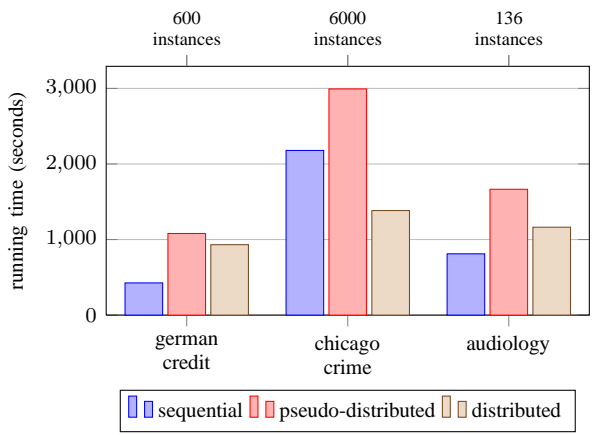
\begin{figure}[htbp]
				\centering
				\input{figures/preliminary_analysis/plot_stats_running_time}
				\caption{Comparison of the running time}
				\label{fig:Comparison of the running time}
			\end{figure}
			
			The results in figure \ref{fig:Comparison of the running time} are rather variegated. With \textit{German Credit} the winner is without doubt the \textit{Sequential} one and it is the same in another case. The framework must face some considerable intrinsic \textit{Hadoop} bottlenecks such as \textit{map} initialisation, data reading, copying and memorisation etc.. Every single made choice adds overhead to the running time and often this is quite enough. The proposed version of the framework does not consider every single aspect of \textit{Hadoop} and this allows the deduction that a further optimisation might increase the performance of the parallel version compared with sequential one.
			
			Nevertheless, the performance of the \textit{distributed} version is not so distant. In one case, with the \textit{Chicago Crime} dataset that is rich of instances, the \textit{distributed} framework version beats the \textit{sequential} one quite a lot. It is predictable that the \textit{distributed} version performs better when the amount of computation is large, because of its inherent distributing nature.
			
			Since the \textit{pseudo-distributed} is always second to the \textit{distributed}, this suggests that is worth subdividing the work among multiple nodes.

%% file: tables/preliminary_analysis/sequential_version.tex
\begin{tabularx}{\linewidth}{XXXXX}
	\toprule
	Type & Arch & CPUs & RAM (GB) & Storage (GB) \\
	\midrule
	m1.large & 64-bit & $2$ & $7.5$ & $2$ x $420$ \\
	\bottomrule
\end{tabularx}

%% file: tables/preliminary_analysis/pseudo-distributed_version.tex
\begin{tabularx}{\linewidth}{lXXXXX}
	\toprule
	Hadoop Role & Type & Arch & CPUs & RAM (GB) & Storage (GB) \\
	\midrule
	All Hadoop roles & m1.large & 64-bit & $2$ & $7.5$ & $2$ x $420$ \\
	\bottomrule
\end{tabularx}

%% file: tables/preliminary_analysis/distributed_version.tex
\begin{tabularx}{\linewidth}{lXXXXX}
	\toprule
	Hadoop Role & Type & Arch & CPUs & RAM (GB) & Storage (GB) \\
	\midrule
	JobTracker, NameNode & m1.large & 64-bit & $2$ & $7.5$ & $2$ x $420$ \\
	TaskTracker, DataNode & m1.large & 64-bit & $2$ & $7.5$ & $2$ x $420$ \\
	TaskTracker, DataNode & m1.large & 64-bit & $2$ & $7.5$ & $2$ x $420$ \\
	TaskTracker, DataNode & m1.large & 64-bit & $2$ & $7.5$ & $2$ x $420$ \\
	TaskTracker, DataNode & m1.large & 64-bit & $2$ & $7.5$ & $2$ x $420$ \\
	\bottomrule
\end{tabularx}

%% file: figures/preliminary_analysis/pie_chart_framework_developing.tex
\begin{tikzpicture}
	\pie[sum=auto, text=legend, radius=1.5, color={blue!30, cyan!30}, style=very thin]{3908/Framework, 1903/Test}
\end{tikzpicture}

%% file: figures/preliminary_analysis/pie_chart_application_developing.tex
\begin{tikzpicture}
	\pie[sum=auto, text=legend, radius=1.5, color={red!30, blue!30}, style=very thin]{535/Application, 3908/Framework}
\end{tikzpicture}

%% file: figures/preliminary_analysis/plot_stats_attributes.tex
\begin{tikzpicture}
	\begin{axis}
		[
			ybar,
			width=0.9\linewidth,
			height=0.55\linewidth,
			bar width=0.5cm,
			ymajorgrids,
			ylabel={\footnotesize attributes},
			y tick label style={font=\footnotesize},
			symbolic x coords={german credit, chicago crime, audiology},
			x tick label style={font=\footnotesize, text width=1.5cm, align=center},
			enlarge x limits=0.25,
			xtick=data,
			legend style={at={(0.5,-0.3)}, anchor=north, legend columns=-1, font=\footnotesize},
			extra x ticks={german credit, chicago crime, audiology},
			extra x tick labels={600 \\ instances, 6000 \\ instances, 136 \\ instances, 203 \\ instances},
			extra x tick style={tick pos=right, ticklabel pos=right, x tick label style={font=\scriptsize}}
		]
		\addplot coordinates {
			(german credit, 8)
			(chicago crime, 6)
			(audiology, 34)
		};
		\addplot coordinates {
			(german credit, 6)
			(chicago crime, 10)
			(audiology, 43)
		};
		\addplot coordinates {
			(german credit, 6)
			(chicago crime, 8)
			(audiology, 41)
		};
		\legend{sequential,pseudo-distributed,distributed}
	\end{axis}
\end{tikzpicture}

%% file: figures/preliminary_analysis/plot_stats_accuracy.tex
\begin{tikzpicture}
	\begin{axis}
		[
			ybar,
			width=0.9\linewidth,
			height=0.55\linewidth,
			bar width=0.5cm,
			ymajorgrids,
			ylabel={\footnotesize accuracy (\%)},	
			y tick label style={font=\footnotesize},
			symbolic x coords={german credit, chicago crime, audiology},
			x tick label style={font=\footnotesize, text width=1.5cm, align=center},
			enlarge x limits=0.25,
			xtick=data,
			legend style={at={(0.5,-0.3)}, anchor=north, legend columns=-1, font=\footnotesize},
			extra x ticks={german credit, chicago crime, audiology},
			extra x tick labels={600 \\ instances, 6000 \\ instances, 136 \\ instances},
			extra x tick style={tick pos=right, ticklabel pos=right, x tick label style={font=\scriptsize}}
		]
		\addplot coordinates {
			(german credit, 69.25)
			(chicago crime, 89.55)
			(audiology, 68.88)
		};
		\addplot coordinates {
			(german credit, 72.00)
			(chicago crime, 89.40)
			(audiology, 71.11)
		};
		\addplot coordinates {
			(german credit, 72.00)
			(chicago crime, 89.45)
			(audiology, 72.22)
		};
		\legend{sequential,pseudo-distributed,distributed}
	\end{axis}
\end{tikzpicture}

%% file: figures/preliminary_analysis/plot_stats_running_time.tex
\begin{tikzpicture}
	\begin{axis}
		[
			ybar,
			width=0.9\linewidth,
			height=0.55\linewidth,
			bar width=0.5cm,
			ymajorgrids,
			ylabel={\footnotesize running time (seconds)},	
			y tick label style={font=\footnotesize},
			ymin=0,
			symbolic x coords={german credit, chicago crime, audiology},
			x tick label style={font=\footnotesize, text width=1.5cm, align=center},
			enlarge x limits=0.25,
			xtick=data,
			legend style={at={(0.5,-0.3)}, anchor=north, legend columns=-1, font=\footnotesize},
			extra x ticks={german credit, chicago crime, audiology},
			extra x tick labels={600 \\ instances, 6000 \\ instances, 136 \\ instances},
			extra x tick style={tick pos=right, ticklabel pos=right, x tick label style={font=\scriptsize}}
		]
		\addplot coordinates {
			(german credit, 426)
			(chicago crime, 2179)
			(audiology, 811)
		};
		\addplot coordinates {
			(german credit, 1080)
			(chicago crime, 2993)
			(audiology, 1665)
		};
		\addplot coordinates {
			(german credit, 932)
			(chicago crime, 1384)
			(audiology, 1163)
		};
		\legend{sequential,pseudo-distributed,distributed}
	\end{axis}
\end{tikzpicture}

%% file: conclusions_and_future_work.tex
\section{Conclusions and Future Work}
\label{sec:Conclusions and Future Work}
	
	The test was preliminary for some reasons:
	\begin{itemize}
		\item Defeating the sequential version of \textit{Weka}, which is specialised in \textit{Machine Learning} algorithms, is a challenge itself;
		\item The probability factor of \textit{Genetic Algorithms} needs an average of a greater number of executions of the test in order to be more reliable.
	\end{itemize}
	
	The results obtained in the example of use of the framework show that writing \textit{GAs} with the framework is effective because of the good ratio of number of lines of code for developing the application to the ones of the framework itself.
	
	Even though the running time is not always on its side, some clues suggest that an \enquote{ad hoc} optimisation of \textit{Hadoop} might improve considerably the performance. The case in which the sequential version is defeated suggests that for large size instances it is worth parallelising.
	
	Nowadays it is possible to rent a cluster for little money without investing on an expensive and dedicate hardware. In a few minutes everything is completely operative so as to run any type of distributed applications. This fact together with the saving of money, the easiness with which it is possible to develop the algorithms and its flexibility, give this framework a relevant meaning. For all these reasons, it can be worth improving it in the future. It can be advantageous to convert the project into an \textit{open-source} project. The current complexity of the framework needs a specific care in the discovery of possible bugs and the potential interest covers the prospective of the existence of a dedicate community.
	
	Even though many of the most common basic functions are already implemented, it could be useful to add some other implementations to better cover the possible needed cases. Moreover, some tools to treat the data read and produced by the framework could help make it more compatible with external applications.
	
	The most important improvement needed is analysing the possible bottlenecks caused by the inner structure of \textit{Hadoop}. As the features of \textit{Hadoop} that can treat data in a more efficient way are many, it could be worth considering them in a future development.

%% file: bibliography.tex
\balance

\printbibliography[heading=bibintoc]